\documentclass[11pt]{article}
\usepackage[utf8]{inputenc}
\usepackage[english]{babel}
\usepackage[font=small,labelfont=bf]{caption}
\usepackage{geometry}
\usepackage[sort&compress]{natbib}
\usepackage{amsmath}
\usepackage{pxfonts}
\usepackage{graphicx}
\usepackage{setspace}
\usepackage{hyperref}
\usepackage{lineno}
\usepackage{booktabs}

\newcommand{\crossentropyContent}{{1}}
\newcommand{\crossentropyFunction}{{2}}
\newcommand{\crossentropyPOS}{{3}}
\newcommand{\ttestsContent}{{4}}
\newcommand{\ttestsFunction}{{5}}
\newcommand{\ttestsPOS}{{6}}
\newcommand{\confusion}{{7}}
\newcommand{\mds}{{8}}

\newcommand{\authortableContent}{{1}}
\newcommand{\authortableFunction}{{2}}
\newcommand{\authortablePOS}{{3}}

\title{A Stylometric Application of Large Language Models}

\author{Harrison F. Stropkay, Jiayi Chen, Mohammad J. Latifi,\\
Daniel N. Rockmore, and Jeremy R. Manning\\
Dartmouth College \\
Hanover, NH 03755, USA \\
\texttt{\{harrison.f.stropkay.25, jiayi.chen.gr, mohammad.javad.latifi.jebelli}\\\texttt{daniel.n.rockmore, jeremy.r.manning\}@dartmouth.edu}}

\begin{document}
\maketitle

\begin{abstract}

We show that large language models (LLMs) can be used to distinguish the
writings of different authors. Specifically, an individual GPT-2 model, trained
from scratch on the works of one author, will predict held-out text from that
author more accurately than held-out text from other authors. We suggest that,
in this way, a model trained on one author's works embodies the unique writing
style of that author. We first demonstrate our approach on books written by
eight different (known) authors. We also use this approach to confirm R. P.
Thompson's authorship of the well-studied 15\textsuperscript{th} book of the
\textit{Oz} series, originally attributed to F. L. Baum.

\end{abstract}

\section{Introduction}

Herein we introduce {\em predictive comparison}, a new LLM-based relative
stylometric measure. It derives from a simple idea, that if an LLM can be
trained to write like---i.e., in the style of---a given author by training on
their work~\citep[e.g., ][]{Mikr25}, then the degree to which such a model can
predict another author's work could be a measure of stylistic similarity. This
approach builds upon a growing body of work applying language models to
authorship attribution~\citep{HuanEtal25,UcheEtal20}, extending established
information-theoretic methods in stylometry~\citep{JuolBaay05,ZhaoEtal06}.

Recent work has demonstrated the effectiveness of using perplexity and
cross-entropy loss from fine-tuned language models for authorship
attribution~\citep{HuanEtal25}, achieving state-of-the-art performance on
standard benchmarks. Unlike traditional stylometric approaches that rely on the
direct articulation of particular features such as function word
frequencies~\citep{MostWall63} or syntactic patterns~\citep{Holm98}, large
language models can capture complex, hierarchical patterns in authorial
style~\citep{FabiEtal20}. This shift from explicit feature engineering to
learned representations parallels broader trends in computational literary
analysis~\citep{More00,UndeEtal19} and digital humanities~\citep{HughEtal12}.

In this paper we show, using a small set of authors and their works, that large
language models capture author-specific writing patterns. Our method differs
from related approaches~\citep{Reza25} in scale (we use entire books rather
than individual sentences) and in our reliance solely on cross-entropy loss as
a measure of stylometric distance. This in turn suggests a notion of
stylometric distance derived from the cross-entropy loss assigned to held-out
texts by models trained on known works of different authors. We believe this
approach could be of use in considering questions of authorial influence and
stylistic evolution~\citep{HughEtal12}. Lastly, this further suggests a
literary attribution tool~\citep[a common use of stylometric techniques;
][]{MostWall63,MostWall84,NiloBino03,Juol08} that would assign an unknown or
contested work to the model (and author) under which predictive comparison
generates the smallest loss. We illustrate this on the well-known attribution
problem of the 15\textsuperscript{th} book in the \emph{Oz} series, confirming
what is now the accepted attribution.

\section{Methods}

In this section, we outline our methodology for identifying stylometric
signatures using large language models. For each selected author, we train a
GPT-2 model~\citep{RadfEtal19} on that author's corpus. We then use the trained
model to compute the cross-entropy loss on held-out texts from both the target
author and each of the other authors in the dataset. By comparing these losses,
we assess whether the model captures author-specific stylistic patterns: a
model trained on a given author should exhibit lower loss when predicting that
author's own texts as compared to the texts of others.

\subsection{Data and preprocessing \label{sec:data}}

We consider a dataset comprising books by eight authors: Jane Austen, L. Frank
Baum, Charles Dickens, F. Scott Fitzgerald, Herman Melville, Rosemary Plumly
Thompson, Mark Twain, and H. G. Wells. We selected these authors because their
writings are well-represented in Project Gutenberg, are all in the public
domain, and are written in English---eliminating any potential confounds due to
translation. For each book, we pre-process the text by stripping Project
Gutenberg metadata, publisher information, illustration tags, transcriber
notes, prefaces, tables of contents, and chapter headings. We standardize
whitespace, remove non-ASCII characters, and lowercase all alphabetic
characters. Basic statistics on token lengths and the full list of books used
are provided in the Appendix.

To construct training data for each author, we randomly select one book to hold
out for evaluation and train their model using the remaining books. To ensure
fair comparisons across authors, we standardize the number of training tokens
per author by truncating each author's corpus. This token budget is determined
by removing the longest book from each author's set and then taking the
smallest of the (remaining) total token counts. For our dataset, this yields a
fixed training token budget of 643,041 tokens.

To construct a truncated corpus of 643,041 tokens for each author, we sample
one contiguous sub-sequence from each book in their training corpus (after
holding out a to-be-evaluated book). The length of the sub-sequence sampled
from book~$i$ is proportional to its original length:
\[
\text{length}_i = 643{,}041 \times \frac{\text{tokens in book
  }i}{\text{total tokens in corpus}}.
\]
The starting position of each sub-sequence is chosen uniformly at random,
ensuring the sample fits within the book's bounds. Finally, we shuffle and then
concatenate the sampled sub-sequences from each book, resulting in a single
643,041-token training sequence for each author. This process is repeated for
each of 10 random seeds, yielding 10 different training corpora for each
author.

\subsection{Model architecture, training, and evaluation}

For each author, we train GPT-2 language models from scratch using the
\texttt{GPT2LMHeadModel} class from the Hugging Face \texttt{Transformers}
library with custom architecture settings: a context window of 1024 tokens, an
embedding dimension of 128, 8 transformer layers, and 8 attention heads per
layer. We fit each model using the AdamW optimizer~\citep{LoshHutt17} with a
learning rate of $5 \times 10^{-5}$ to minimize the cross-entropy loss on the
training data. We train models using a causal language modeling objective,
whereby the model iteratively predicts the next token in the sequence given all
of the previous tokens in the same training sequence.

We construct training samples by sampling 1024-token chunks from the truncated
corpus for the given author and random seed (constructed as described above,
using contiguous sub-sequences selected from all but one of their books). Each
training epoch consists of 40 batches, each containing 16 sequences of 1024
tokens. This results in a total of 655,360 tokens per epoch. We continue
training until the cross-entropy loss falls to 3.0 or lower. (We decided on
this threshold after taking random draws from the models trained on Baum's and
Thompson's \textit{Oz} books and manually inspecting the quality of the
resulting samples.) Training to a fixed loss threshold (e.g., as opposed to
training for a fixed number of epochs) enables us to fairly compare model
performance across authors, which is the central component of our stylometric
analyses.

We evaluate the models using the held-out book from the corresponding author.
We partition the held-out book into 1024-token chunks to ensure that each token
in the evaluation set contributes equally to the computed loss. We repeat the
full process (of selecting a held-out book at random and training the model
using randomly selected samples from the remaining books) using 10 different
random seeds. This approach enables us to assess the robustness of our results
and to ensure that the models are not overfitting to a specific book or random
sample.

\subsection{Investigating the contributions of function words, content
words, and parts of speech \label{subsec:ablation}}

In order to investigate the contributions of different types of words to the
stylometric signatures captured by our models, we carried out additional
analyses using modified corpora. First, we created content-word-only corpora by
replacing all function words with a special token, \texttt{<FUNC>}. Function
words were identified using scikit-learn's list of English stop
words~\citep{PedrEtal11}. Next, we created function-word-only corpora by
replacing all content (i.e., non-function) words with a \texttt{<CONTENT>}
token. Finally, we created part-of-speech-only corpora by using the Natural
Language Toolkit~\citep[NLTK; ][]{BirdLope04} to replace each word with its
corresponding part-of-speech tag. We then re-trained our models on each of
these modified corpora, following the same methodology as described above.

\begin{figure*}[t]
  \centering
  \includegraphics[width=\textwidth]{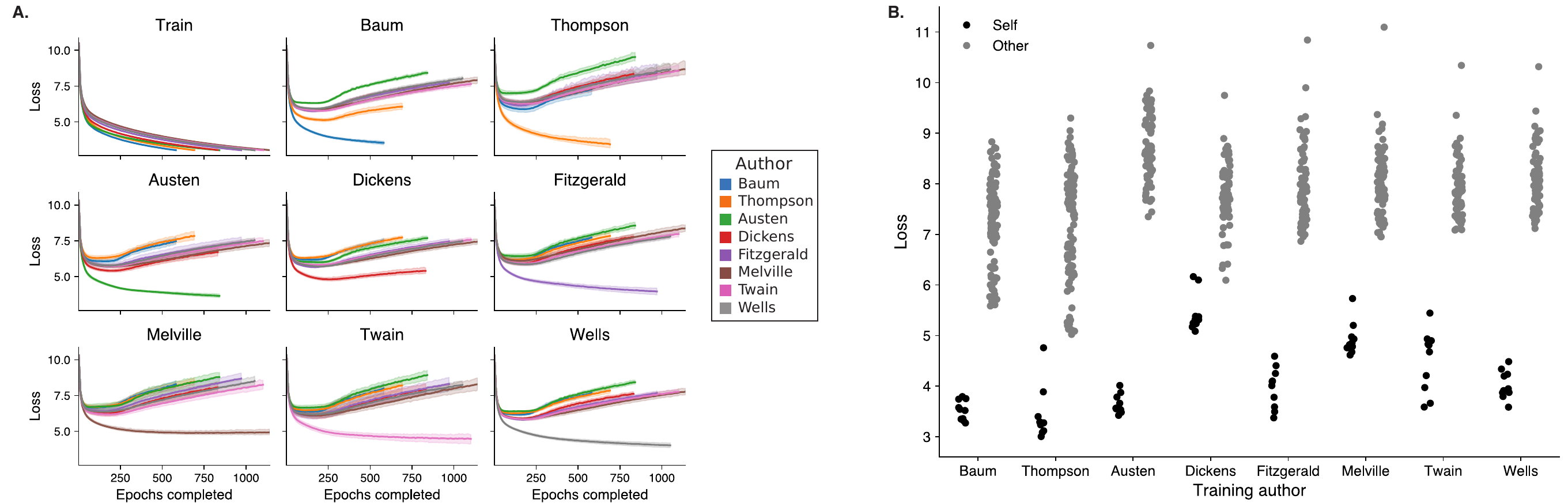}

\caption{\textbf{Cross-entropy loss across models and authors.} \textbf{A.}
Average cross-entropy loss on \textit{Train}ing data and held-out test data
from each author, plotted as a function of the number of training epochs. Each
color denotes a model trained on a single author's work. Error ribbons denote
bootstrap-estimated 95\% confidence intervals over 10 random seeds. \textbf{B.}
Cross-entropy loss assigned to held-out test data by each author's model
($x$-axis). Held-out test data is either from the \textit{same} author (black)
or from \textit{other} authors (gray). Each dot denotes the average loss
(across all 1024-token chunks) for a single random seed. See Supplementary
Materials for analogous plots using models trained on only content words (Supp.
Fig.~\crossentropyContent), only function words (Supp.
Fig.~\crossentropyFunction), and only parts of speech (Supp.
Fig.~\crossentropyPOS).}
\label{fig:all-losses}
\end{figure*}

\section{Results}

\subsection{Predictive comparison testing of eight classic authors}

We carried out predictive comparison testing on eight classic authors (see
Sec.~\ref{sec:data}). The top-left sub-panel of Figure~\ref{fig:all-losses}A
(labeled ``Train'') shows the average training loss for each author's model,
computed over 10 random seeds. Training losses are comparable across models,
indicating that the models are trained to similar levels of performance. The
other sub-panels of Figure~\ref{fig:all-losses}A show the average predictive
(cross-entropy) loss, for each author's model, on held-out texts from each
author. For every author's held-out text, the model trained on the same
author's writings produces the lowest loss, indicating a clear preference for
its own author's stylistic patterns. As shown in Figure~\ref{fig:all-losses}B,
across every author we considered, and for every random seed, models trained
and tested on the same author always yield smaller losses than models trained
on one author and tested on another. Indeed, we achieve perfect (100\%)
classification accuracy when matching authors with held-out texts by labeling
the held-out text according to which model produces the smallest loss.

\begin{figure*}[t]
  \centering
  \includegraphics[width=\textwidth]{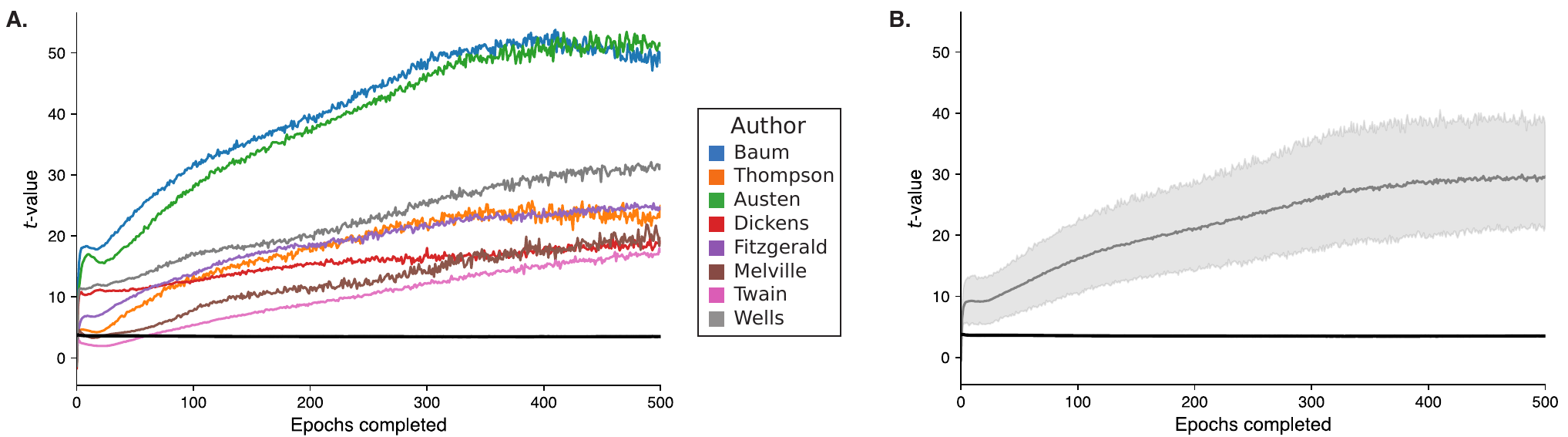}

\caption{\textbf{Same vs. other author comparisons, by model.} \textbf{A.} Each
curve denotes, as a function of the number of training epochs, the the
$t$-statistic from a $t$-test comparing the distribution of losses (across
random seeds) assigned to held-out texts from the given author (color) versus
held-out texts from all other authors. \textbf{B.} The average $t$-statistic
across all eight authors, as a function of the number of training epochs. The
black curves in both panels indicates the average $t$-value corresponding to $p
= 0.001$, for each epoch. Error ribbons denote bootstrap-estimated 95\%
confidence intervals across authors. See Supplementary Materials for analogous
plots using models trained on only content words (Supp. Fig.~\ttestsContent),
only function words (Supp. Fig.~\ttestsFunction), and only parts of speech
(Supp. Fig.~\ttestsPOS).} 

\label{fig:t-stats} 
\end{figure*}

We also wondered how many training epochs were required for the models to
reliably distinguish author styles. We compared the distributions (across
random seeds) of average cross-entropy losses for each author's model computed
for held-out text from the \textit{same} author versus for held-out text from
\textit{other} authors. Figure~\ref{fig:t-stats}A displays the $t$-values from
$t$-tests comparing these same versus other loss distributions for each of the
first 500 training epochs. For all authors except Twain, the $t$-tests yielded
$p$-values below $0.001$ after just one or two epochs, indicating that the models
rapidly acquire author-specific stylometric patterns. For Twain, this threshold
is crossed at epoch 77. Figure~\ref{fig:t-stats}B shows the average $t$-values
across all eight authors as a function of the number of training epochs (final
epoch: $t(9) = 13.196, p = 3.41 \times 10^{-7}$). This latter plot provides an
estimate of the performance we might expect to see in the general case (e.g.,
across a larger set of authors). Table~\ref{tab:t-tests} summarizes the results
of the $t$-tests for each author's model after training is complete.

\begin{table}[h]
\centering
\small
\begin{tabular}{lccc}
\hline
\textbf{Model} & \textbf{$t$-stat} & \textbf{df} & \textbf{$p$-value}\\
\hline
Baum        & 48.39 & 31.53 & $3.69 \times 10^{-31}$  \\
Thompson    & 22.35 & 16.39 & $1.04 \times 10^{-13}$ \\
Austen      & 50.64 & 47.38 & $6.48 \times 10^{-43}$ \\
Dickens     & 16.37 & 17.84 & $3.46 \times 10^{-12}$ \\
Fitzgerald  & 25.94 & 23.13 & $1.55 \times 10^{-16}$ \\
Melville    & 23.38 & 23.13 & $1.35 \times 10^{-17}$ \\
Twain       & 16.74 & 11.27 & $2.60 \times 10^{-9}$ \\
Wells       & 35.73 & 23.68 & $4.15 \times 10^{-22}$ \\
\hline
\end{tabular}

\caption{\textbf{Loss differences between same-author and other-author texts.}
Each row displays the results of a $t$-test comparing the average loss values
assigned by each author's model (after training is complete) to the author's
held-out text and to the other authors' randomly sampled texts. See
Supplementary Materials for analogous tables using models trained on only
content words (Supp. Tab.~\authortableContent), only function words (Supp.
Tab.~\authortableFunction), and only parts of speech (Supp.
Tab.~\authortablePOS).}

\label{tab:t-tests}
\end{table}

\begin{figure*}[t]
  \centering
  \includegraphics[width=0.6\textwidth]{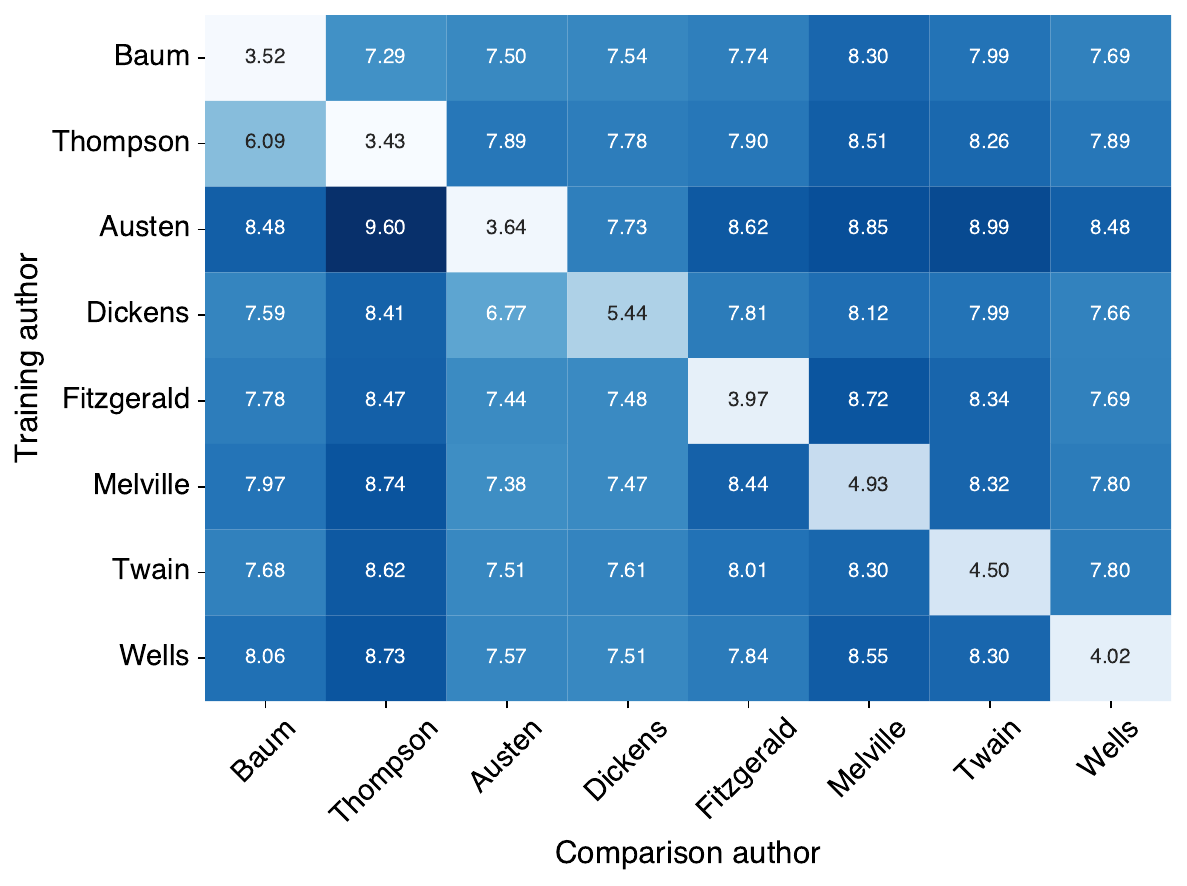}

  \caption{\textbf{Confusion matrix.} The matrix displays the average cross-entropy loss assigned by models trained on each
 author's writing (column) to held-out texts from each author (row), after subtracting
 the native author's baseline loss.  See Supplementary Materials for analogous plots using models trained on only content words, function words, and parts of speech (Supp. Fig.~\confusion).}
\label{fig:confusion-matrix}
\end{figure*}

Despite achieving perfect classification accuracy, not all authors are equally
distinctive. For example, we reasoned that authors with similar writing styles
might be more confusable (i.e., yielding relatively smaller losses for models
trained across different authors). We computed the average loss for each author
using the models trained on the other authors' texts
(Fig.~\ref{fig:confusion-matrix}). Authors with similar writing styles (e.g.,
Baum and Thompson) yield relatively small losses when evaluated using models
trained on the other author's texts. In contrast, authors with more distinct
writing styles (e.g., Austen and Thompson) yield relatively large losses when
evaluated using each other's models. To illustrate these patterns, we also
project the losses into a 3D space using multidimensional scaling~\citep[MDS;
][]{Krus64} applied to the pairwise correlations between rows of the loss
matrix, excluding the diagonal entries (i.e., the losses obtained using each
author's model when applied to their own held-out text; Fig.~\ref{fig:mds}). We
suggest that this approach might lend itself to further exploration and
consideration by literature scholars, particularly if extended to a larger
embedding space. For the purposes of our present work, however, we provide the
plot solely as a provocative demonstration.

\subsection{Stylometric distance~\label{sec:distance}}

\begin{figure*}[t]
  \centering
  \includegraphics[width=0.6\textwidth]{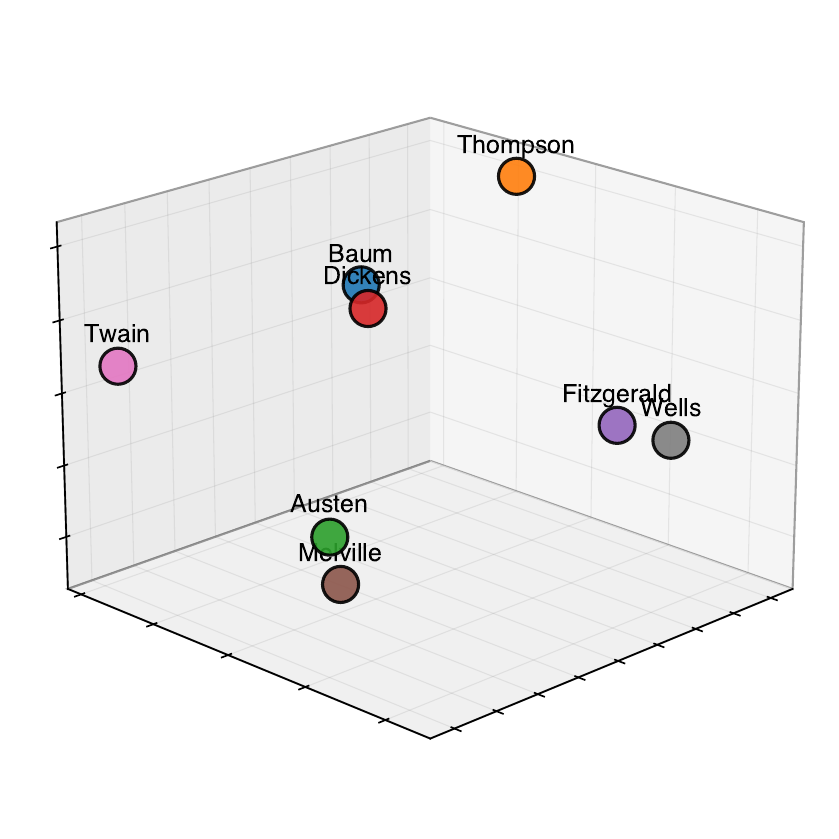}

  \caption{\textbf{Multidimensional scaling plot.} Three-dimensional
    MDS projection of the (symmetrized) average cross-entropy loss matrix shown in
    Figure~\ref{fig:confusion-matrix}. See Supplementary Materials for analogous plots using models trained on only content words, function words, and parts of speech (Supp. Fig.~\mds).}
\label{fig:mds}
\end{figure*}

As indicated by Figure~\ref{fig:mds}, predictive comparison suggests a
natural notion of distance between authorial styles. Let $L_j(i)$ denote the
average loss of a work of author $i$ for a model trained on author $j$ (entry
$i,j$ of the average loss matrix in Fig.~\ref{fig:confusion-matrix}). Let
$\overline{L_j(i)} = L_j(i)-L_j(j)$, normalizing the entries by subtracting the
native author's baseline loss. Then define the LLM-based {\em stylometric
distance}, $d(i,j) = \frac{1}{2}\left(\overline{L_j(i)} +
\overline{L_i(j)}\right)$. Thus, Figure~\ref{fig:mds} is a visualization of
the relative ``distances" among our author set.

\subsection{Predictive attribution of the 15\textsuperscript{th} \emph{Oz} book}

\begin{figure*}[t]
  \centering
  \includegraphics[width=0.8\textwidth]{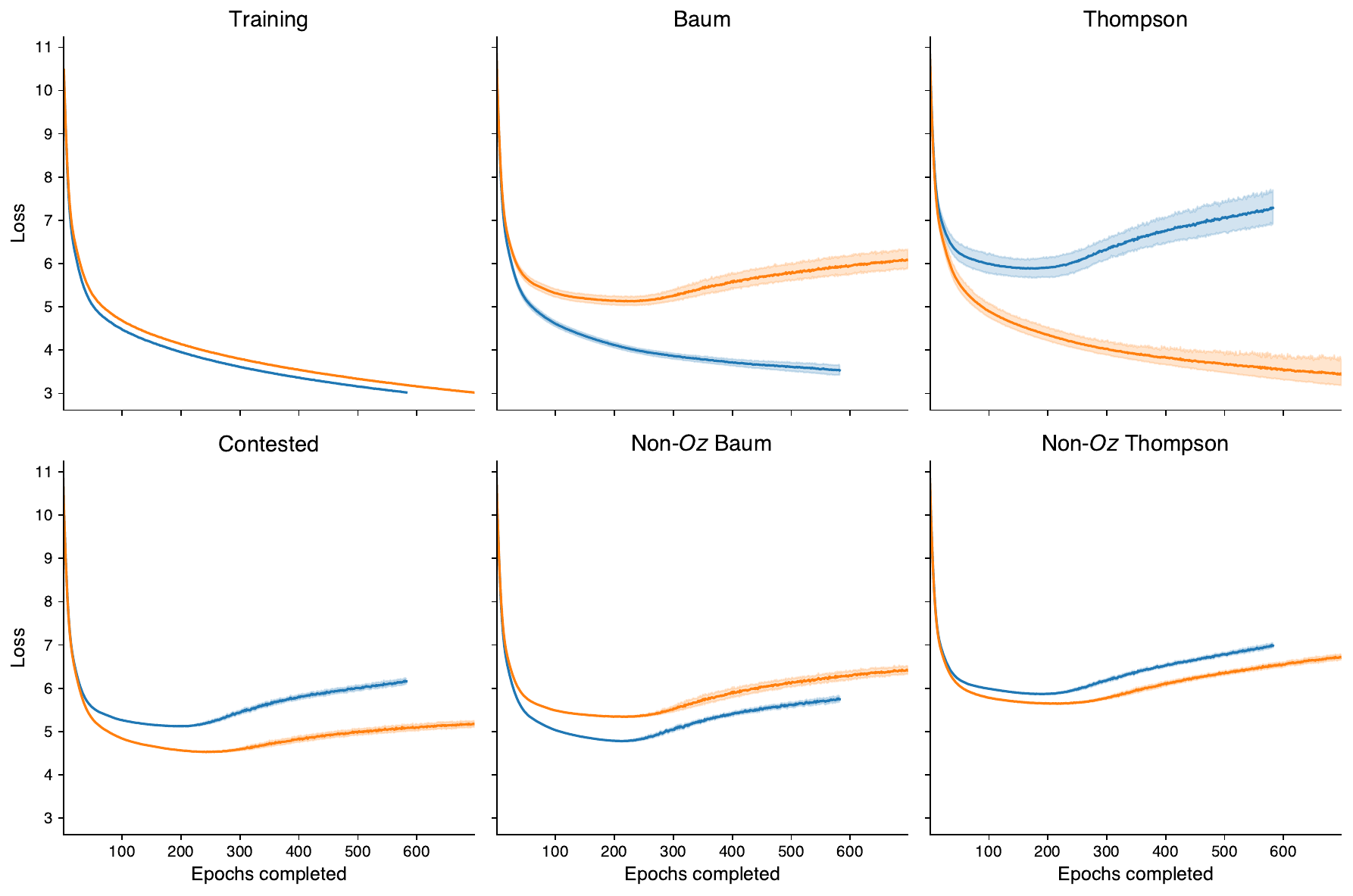}

\caption{\textbf{Cross-entropy loss across models and \textit{Oz} authors.} The
top sub-panels replicate the Baum (blue) and Thompson (orange) results from
Figure~\ref{fig:all-losses}---i.e., that a given Thompson is well-distinguished
from Baum and vice-versa (the two rightmost top sub-panels; error ribbons
denote bootstrap-estimated 95\% confidence intervals over 10 random seeds). The
bottom sub-panels show the cross-entropy loss assigned to a held-out text whose
authorship is contested (lower left), to a held-out non-\textit{Oz} text by
Baum (lower center), and to a held-out non-\textit{Oz} text by Thompson (lower
right). I.e., the contested book shows lower loss for Thompson-trained models;
a non-Oz Baum book shows lower loss for Baum-trained models; and a non-Oz
Thompson book shows lower loss for Thompson-trained models.}

\label{fig:oz-losses}
\end{figure*}

Attribution is another application of predictive comparison. We illustrate with
the well-known example of the contested authorship of the
15\textsuperscript{th} \emph{Oz} book (in a thirty-one book series), widely
believed to have been written by Ruth Plumly Thompson, but originally
attributed to L. Frank Baum~\citep{NiloBino03}. We applied predictive
comparison to the 15\textsuperscript{th} \emph{Oz} book, using models trained
on Baum and Thompson's undisputed \emph{Oz} books. As shown in the bottom left
sub-panel of Figure~\ref{fig:oz-losses}, we find lower loss for the
Thompson-trained model than for the Baum-trained model, indicating that the
contested book is indeed more similar to Thompson's writing style than to
Baum's. We also applied both models to a non-\emph{Oz} book by Baum (bottom
center) and Thompson (bottom right). We see lower losses for the correct author
in each case, demonstrating that predictive comparison is robust to thematic
differences within the same author's writings.

\subsection{Ablation studies: content words, function words, and parts of speech}

The above analyses show that LLMs trained on one author's works can effectively
capture the distinctive statistical patterns of that author's writing style. We
carried out a series of ablation studies to investigate the contributions of
different aspects of writing style. Specifically, we constructed three modified
corpora for each author: (1) content-word-only corpora, in which all function
words were replaced with a special token; (2) function-word-only corpora, in
which all content words were replaced with a special token; and (3)
part-of-speech-only corpora, in which each word was replaced with its
corresponding part-of-speech tag (see \nameref{subsec:ablation}). We then
re-trained our models on each of these modified corpora and repeated the
predictive comparison analyses (Supp. Figs.~\crossentropyContent--\ttestsPOS).

The models trained on the content-word-only corpora were intended to capture
stylistic patterns related to vocabulary choice and thematic content. The
models trained on the function-word-only corpora were intended to capture
syntactic and grammatical patterns that transcended story-specific content.
Finally, the models trained on the part-of-speech-only corpora were intended to
capture higher-level syntactic patterns while abstracting away from specific
word choices. Models trained on a single author's texts from each of these
modified corpora all converged, achieving training losses below 3.0 well within
500 training epochs (Supp. Figs.~\crossentropyContent, \crossentropyFunction,
and \crossentropyPOS).

We found that models trained on content-word-only corpora reliably learned
author-specific patterns for 6 of the 8 authors (Supp.
Figs.~\crossentropyContent~and~\ttestsContent, Supp.
Tab.~\authortableContent). Overall, by the final training epoch, the average
$t$-values across all models and held-out texts were reliably greater than zero
($t(9) = 8.438, p = 1.44 \times 10^{-5}$). However, models trained only on
content words were significantly less effective at distinguishing authors than
models trained on the intact texts ($t(11.77) = 3.21, p = 7.68 \times
10^{-3}$).

Models trained on function-word-only corpora reliably learned author-specific
patterns for 5 of the 8 authors (Supp.
Figs.~\crossentropyFunction~and~\ttestsFunction, Supp.
Tab.~\authortableFunction). Overall, by the final training epoch, the average
$t$-values across all models and held-out texts were reliably greater than zero
($t(9) = 4.428, p = 1.65 \times 10^{-3}$). These models were also significantly
less effective at distinguishing authors than models trained on the intact
texts ($t(8.36) = 4.82, p = 1.15 \times 10^{-3}$), but not significantly
different from models trained on content-word-only corpora ($t(10.29) = 1.81, p
= 0.100$).

Models trained on part-of-speech-only corpora reliably learned author-specific
patterns for just 3 of the 8 authors (Supp.
Figs.~\crossentropyPOS~and~\ttestsPOS, Supp. Tab.~\authortablePOS). By the
final training epoch, the average $t$-values across all models and held-out
texts were not reliably greater than zero ($t(9) = 1.616, p = 0.141$). These
models were significantly less effective at distinguishing authors than models
trained on the intact texts ($t(7.36) = 5.72, p = 6.01 \times 10^{-4}$), models
trained on content-word-only corpora ($t(7.90) = 3.10, p = 1.49 \times
10^{-2}$), and models trained on function-word-only corpora ($t(10.41) = 2.11,
p = 6.04 \times 10^{-2}$).

Taken together, these ablation results suggest that both content words and
function words contribute to the author-unique stylometric signatures captured
by our models. In contrast, grammatical structure alone, as reflected in
part-of-speech sequences and captured by our methodology, appears to be more
similar across authors. Distinctiveness notwithstanding, however, models
trained on \textit{all} of the corpora (intact texts, content-word-only,
function-word-only, and part-of-speech-only) all rapidly converged to low
training and evaluation losses. This indicates that all four corpora contain
sufficient statistical regularities for GPT-2 models to learn to reliably
and accurately make next-token predictions.

\section{Discussion}

We introduced predictive comparison, a method for stylometric analysis that
leverages the predictive capabilities of language models trained on individual
authors' works. Our approach rests on a straightforward principle: if a
language model can learn to generate text in an author's style, then the
cross-entropy loss of that model on held-out text should reflect stylistic
similarity. By training separate GPT-2 models for each author and comparing
their predictive performance, we aimed to develop both a measure of stylometric
distance and a practical tool for authorship attribution.

Our results demonstrate the effectiveness of this approach across multiple
dimensions. Models trained and tested on the same author consistently yielded
lower cross-entropy losses than models trained on different authors, achieving
perfect classification accuracy across all eight authors examined. This
separation emerged rapidly during training: for seven of eight authors,
statistically significant discrimination was achieved after just two training
epochs. The resulting stylometric distances proved meaningful, clustering
authors with known stylistic similarities (e.g., Baum and Thompson) while
maintaining clear separation between all author pairs. Finally, our method
successfully resolved the well-studied attribution problem of the
15\textsuperscript{th} \emph{Oz} book, confirming Thompson's authorship in
agreement with traditional stylometric analyses~\citep{NiloBino03}.

We also conducted ablation studies to investigate the contributions of different
aspects of writing style. Models trained on content-word-only and function-word-only
corpora both captured author-specific patterns, though with reduced effectiveness
compared to models trained on intact texts. In contrast, models trained solely on
part-of-speech sequences struggled to distinguish authors, suggesting that
grammatical structure alone is less distinctive. These findings highlight the
importance of both lexical choice and syntactic patterns in shaping authorial style.

\subsection{Relationship to prior work}

Our predictive comparison approach relates closely to recent work using
language model perplexity for authorship attribution~\citep{HuanEtal25}, which
independently developed a similar methodology using fine-tuned (rather than
trained-from-scratch) GPT-2 models. Both approaches exploit the relationship
between perplexity and cross-entropy loss, treating authorship attribution as a
language modeling problem rather than a classification task. This convergent
development suggests that predictive modeling may be a natural framework for
capturing authorial style.

The information-theoretic foundations of our approach connect to earlier work
using cross-entropy~\citep{JuolBaay05} and relative entropy~\citep{ZhaoEtal06}
for stylometry. These methods recognized that authorial style manifests not
just in feature frequencies but in their sequential dependencies---precisely
what language models are designed to capture. Our contribution extends this
line of reasoning to large language models, which can learn these dependencies
implicitly rather than requiring explicit feature engineering.

Compared to classification-based approaches using BERT~\citep{FabiEtal20} or
other transformers~\citep{UcheEtal20}, predictive comparison offers conceptual
simplicity: rather than training a single classifier to distinguish multiple
authors, we train author-specific models that embody each writer's style. This
approach naturally extends to open-set attribution problems where new authors
may be introduced without retraining existing models. However, classification
approaches may be more computationally efficient when dealing with fixed author
sets, as they require training only a single model.

Our reliance on books as training data contrasts with most contemporary
stylometry research, which typically uses shorter texts to enable larger author
sets~\citep{TyoEtal22}. While this limits our experimental scope, it ensures
that our models capture sustained stylistic patterns rather than
topic-specific or context-dependent features that might dominate shorter
texts~\citep{FincBosc24}. The success on full-length books suggests that
predictive comparison can leverage the rich stylistic signal present in
longer texts.

\subsection{Limitations and challenges}

Several limitations constrain the interpretation and application of our results.
The most immediate is the limited experimental scope; we examined only eight
authors writing in English during overlapping historical periods. Whether
predictive comparison maintains its effectiveness across larger author sets,
different languages, or more diverse time periods remains an open question.
The computational requirements of training separate models for each author may
become prohibitive for attribution problems involving hundreds or thousands of
candidate authors.

The opacity of large language models also presents interpretability
challenges~\citep{SchuEtal20}. While our method successfully discriminates
between authors, understanding which stylistic features drive this
discrimination remains elusive. Unlike traditional stylometry, where specific
features (e.g., function word frequencies) can be examined directly, the
distributed representations learned by GPT-2 resist straightforward
interpretation. This ``black box'' nature may limit adoption in domains where
explanations for attribution decisions are required.

Cross-domain robustness represents another significant challenge. Prior work
has shown that language model-based authorship attribution methods can struggle
when training and test texts come from different genres or
topics~\citep{BarlStam20}. Our experiments used books from the same genre for
each author, leaving cross-domain performance unexplored. The strong
performance on Baum and Thompson's \emph{Oz} books versus their non-\emph{Oz}
works provides encouraging evidence, but systematic evaluation across diverse
domains is needed.

The vulnerability of language model-based methods to adversarial
attacks~\citep{QuirMaie19} raises concerns about the reliability of predictive
comparison in adversarial settings. Authors attempting to disguise their style
or imitate others might fool language model-based attribution more easily than
traditional methods that rely on subtler stylistic habits that are difficult to
intentionally emulate. Evaluating robustness against both intentional
obfuscation and unintentional style drift (e.g., authorial development over
time) will be crucial for practical applications.

\subsection{Future directions}

Several research directions could address current limitations while extending
the theoretical and practical reach of predictive comparison. Understanding the
theoretical relationship between cross-entropy loss and stylistic similarity
would provide principled foundations for the approach. Why does minimizing
cross-entropy during training lead to models that capture author-specific
rather than general linguistic patterns? Connecting language model objectives
to stylometric theory could yield insights for both fields.

Developing hybrid approaches that combine predictive comparison with
traditional stylometric features or classification-based language-modeling
methods might offset individual weaknesses. For instance, using cross-entropy
loss as one feature among many in an ensemble model could improve robustness
while maintaining interpretability through traditional features. Alternatively,
predictive comparison could provide initial attributions that are refined using
more interpretable methods.

The scalability challenge invites algorithmic innovations. Rather than training
separate models from scratch for each author, could we use parameter-efficient
fine-tuning methods~\citep{HoulEtal19} to adapt a single base model? Could
authors be represented as vectors in a learned embedding space, with a single
model conditioned on these embeddings? Such approaches might enable attribution
among thousands of authors while maintaining the conceptual advantages of
predictive modeling.

Finally, exploring applications beyond attribution could demonstrate the
broader utility of modeling individual writing styles. For example,
author-specific language models might be used to assist in literary analysis by
generating counterfactual texts, such as what Austen might have written about
modern themes (e.g., the impact of social media on relationships). These
approaches might also help to identify stylistic development within an author's
career, or trace influence networks among authors. These applications would
position predictive comparison within the broader landscape of computational
literary studies.

\subsection{Concluding remarks}

Just as prior work has shown that it is possible to train LLMs to
\textit{write} in the ``style'' or ``voice'' of a given author~\citep[see
e.g.,][]{Mikr25}, our work shows that LLMs may also be used to predict
authorship and measure the stylistic distances between different authors. The
predictive comparison method we have introduced offers a conceptually
straightforward approach: models trained on individual authors' works embody
their unique stylistic patterns, and the cross-entropy loss of these models on
new texts provides a natural measure of stylistic similarity.

The strong empirical results---perfect attribution accuracy and meaningful
stylometric distances---suggest that language models capture robust stylistic
signatures, even when trained on relatively limited data. The convergence of
our approach with concurrent work~\citep{HuanEtal25,Reza25} indicates that the
field may be moving toward predictive modeling as a unifying framework for
computational stylometry. Our finding that models trained only on content words
or function words can still capture author-specific patterns highlights the
multifaceted nature of writing style. Both lexical choice and syntactic
patterns appear to contribute to unique authorial signatures. In contrast,
models trained solely on part-of-speech sequences struggled to differentiate
between most authors, suggesting that grammatical structure alone is less
distinctive. Overall, we suggest that our approach holds promise as a new
technique for machine reading approaches to text-based
disciplines~\citep{More17,More00,Holm98} and the practices of cultural
analytics~\citep{UndeEtal13}.

\section*{Acknowledgments}

We acknowledge helpful discussions with Jacob Bacus, Hung-Tu Chen, and Paxton
Fitzpatrick. This research was supported in part by National Science Foundation
Grant 2145172 to JRM and by a GPU cluster generously donated by the estate of
Daniel J. Milstein.

\section*{Data and code availability}

All code and data needed to reproduce the results in this paper are available
at \url{https://github.com/ContextLab/llm-stylometry}.



\bibliographystyle{apalike}
\bibliography{custom}


\newpage
\appendix

\begin{center}
{\Large \bfseries Appendix: Authors, books, and tokens \par \label{sec:appendix}}
\vspace{1em}
\begin{tabular}{@{}ll|ll@{}}
\toprule
\textbf{Charles Dickens} & \textbf{Tokens} & \textbf{Herman Melville} & \textbf{Tokens} \\
\midrule
A Christmas Carol & 38,906 & I and My Chimney & 15,341 \\
Oliver Twist & 216,100 & Bartleby, the Scrivener & 19,112 \\
The Old Curiosity Shop & 285,895 & Israel Potter & 88,570 \\
Bleak House & 471,630 & Omoo & 134,628 \\
Dombey and Son & 482,161 & Mardi, Vol. II & 150,347 \\
David Copperfield & 479,387 & The Confidence-Man & 129,059 \\
A Tale of Two Cities & 181,593 & White Jacket & 190,577 \\
Nicholas Nickleby & 446,457 & Mardi, Vol. I & 132,358 \\
American Notes & 129,214 & Moby-Dick & 285,066 \\
The Pickwick Papers & 432,546 & Typee & 114,239 \\
Great Expectations & 244,897 & & \\
Martin Chuzzlewit & 455,995 & & \\
Little Dorrit & 449,230 & & \\
Hard Times & 142,759 & & \\
\textbf{Total} & \textbf{4,456,770} & \textbf{Total} & \textbf{1,259,297} \\
\bottomrule
\end{tabular}

\newpage

\begin{tabular}{@{}ll|ll@{}}
\toprule
\textbf{L. Frank Baum} & \textbf{Tokens} & \textbf{Ruth Plumly Thompson} & \textbf{Tokens} \\
\midrule
Ozma of Oz & 52,039 & The Giant Horse of Oz & 51,036 \\
Dorothy and the Wizard in Oz & 53,849 & The Cowardly Lion of Oz & 61,666 \\
Tik-Tok of Oz & 63,781 & Handy Mandy in Oz & 44,778 \\
The Road to Oz & 52,866 & The Gnome King of Oz & 51,687 \\
The Magic of Oz & 51,166 & Grampa in Oz & 55,169 \\
The Patchwork Girl of Oz & 75,703 & Captain Salt in Oz & 61,797 \\
The Wonderful Wizard of Oz & 49,686 & Ozoplaning with the Wizard of Oz & 50,660 \\
The Lost Princess of Oz & 60,418 & The Wishing Horse of Oz & 59,490 \\
The Emerald City of Oz & 70,781 & The Lost King of Oz & 58,105 \\
The Tin Woodman of Oz & 57,338 & The Hungry Tiger of Oz & 53,543 \\
Rinkitink in Oz & 62,241 & The Silver Princess in Oz & 47,964 \\
The Marvelous Land of Oz & 54,733 & Kabumpo in Oz & 62,693 \\
Glinda of Oz & 51,218 & Jack Pumpkinhead of Oz & 49,661 \\
The Scarecrow of Oz & 59,593 & & \\
\textbf{Total} & \textbf{815,412} & \textbf{Total} & \textbf{708,249} \\
\bottomrule
\end{tabular}

\end{center}

\newpage

\begin{center}
\begin{tabular}{@{}ll|ll@{}}
\toprule
\textbf{Jane Austen} & \textbf{Tokens} & \textbf{Mark Twain} & \textbf{Tokens} \\
\midrule
Sense And Sensibility & 153,718 & Adventures Of Huckleberry Finn & 147,655 \\
Mansfield Park & 201,611 & A Connecticut Yankee In King Arthur'S Court & 150,327 \\
Lady Susan & 29,043 & Roughing It & 208,545 \\
Northanger Abbey & 98,090 & The Innocents Abroad & 246,321 \\
Emma & 207,830 & The Adventures Of Tom Sawyer, Complete & 95,059 \\
Pride And Prejudice & 157,777 & The Prince And The Pauper & 88,409 \\
Persuasion & 106,027 & & \\
\textbf{Total} & \textbf{954,096} & \textbf{Total} & \textbf{936,316} \\
\bottomrule
\end{tabular}

\newpage

\begin{tabular}{@{}ll|ll@{}}
\toprule
\textbf{F. Scott Fitzgerald} & \textbf{Tokens} & \textbf{H. G. Wells} & \textbf{Tokens} \\
\midrule
The Beautiful And Damned & 168,147 & The Red Room & 4,944 \\
Flappers And Philosophers & 84,707 & The First Men In The Moon & 87,615 \\
This Side Of Paradise & 100,796 & The Island Of Doctor Moreau & 55,967 \\
All The Sad Young Men & 85,411 & The Open Conspiracy & 40,271 \\
Tales Of The Jazz Age & 109,997 & A Modern Utopia & 105,810 \\
The Pat Hobby Stories & 51,069 & The Sleeper Awakes & 98,228 \\
The Great Gatsby & 65,136 & The New Machiavelli & 185,158 \\
Tender Is The Night & 145,925 & The War Of The Worlds & 75,727 \\
& & Tales Of Space And Time & 94,711 \\
& & The Invisible Man: A Grotesque Romance & 65,584 \\
& & The Time Machine & 40,184 \\
& & The World Set Free & 80,518 \\
\textbf{Total} & \textbf{811,188} & \textbf{Total} & \textbf{934,717} \\
\bottomrule
\end{tabular}
\end{center}

\end{document}


\renewcommand{\figurename}{Supplementary Figure}
\renewcommand{\tablename}{Supplementary Table}

\newcommand{\crossentropy}{{1}}
\newcommand{\ttests}{{2}}
\newcommand{\confusion}{{3}}
\newcommand{\mds}{{4}}

\newcommand{\authortable}{{1}}

\setcounter{equation}{0}
\setcounter{figure}{0}
\setcounter{table}{0}
\setcounter{page}{1}
\setcounter{section}{0}
\makeatletter

\begin{titlepage}
\maketitle
\end{titlepage}

\begin{figure*}[h]
  \centering
  \includegraphics[width=\textwidth]{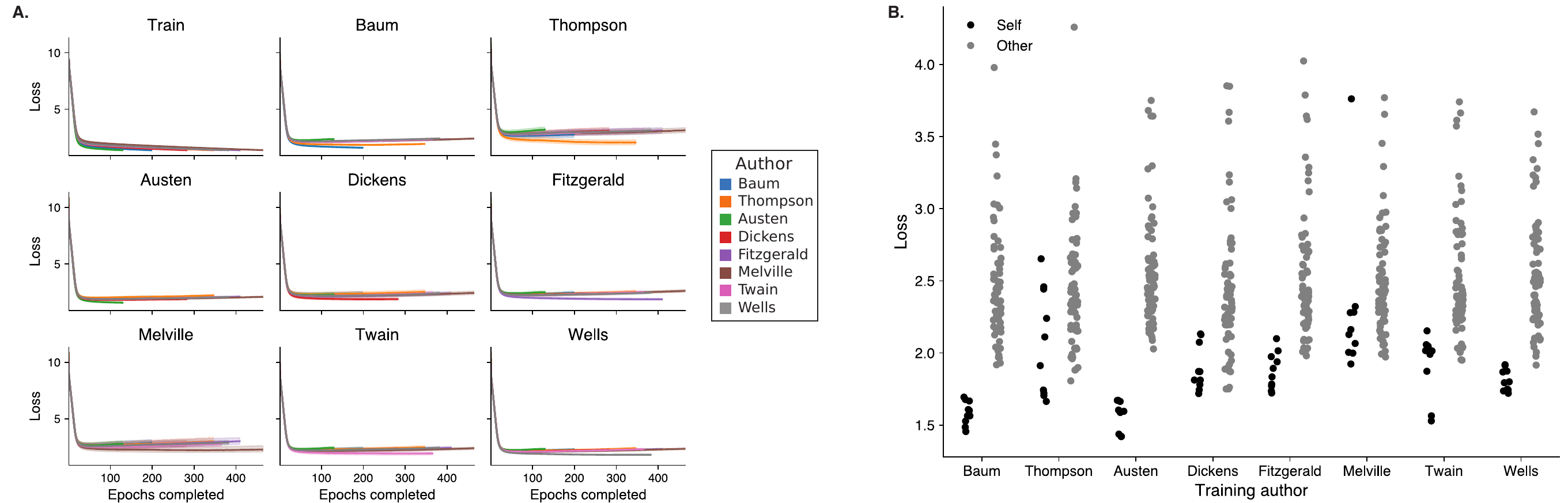}

  \caption{\textbf{Cross-entropy loss across models and
      authors using only content words.} Follows the general format of Figure~\crossentropy~in the main text,
      but uses models trained on only content words. All function words are masked out using \texttt{<FUNC>}.
      \textbf{A.} Average cross-entropy loss on
    \textit{Train}ing data and held-out test data from each author,
    plotted as a function of the number of training epochs. Each color
    denotes a model trained on a single author's work.  Error ribbons
    denote bootstrap-estimated 95\% confidence intervals over 10
    random seeds. \textbf{B.} Cross-entropy loss assigned to held-out
    test data by each author's model ($x$-axis). Held-out test data is
    either from the \textit{same} author (black) or from
    \textit{other} authors (gray). Each dot denotes the average loss
    (across all 1024-token chunks) for a single random seed.}
\label{fig:all-losses-content}
\end{figure*}

\begin{figure*}[h]
  \centering
  \includegraphics[width=\textwidth]{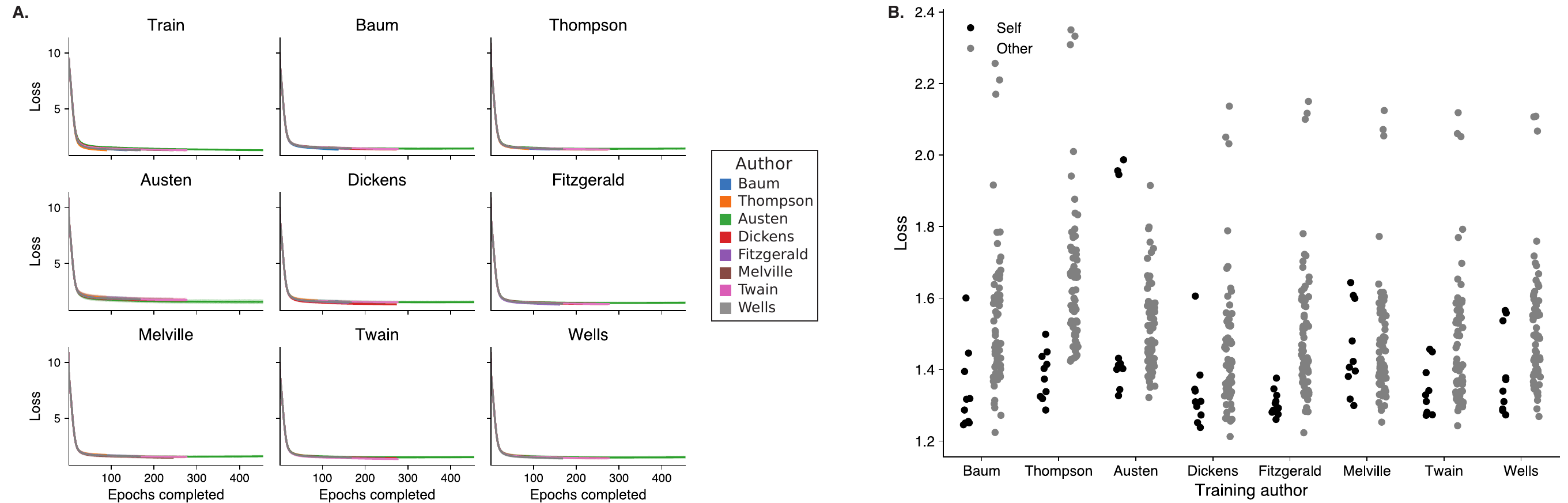}

  \caption{\textbf{Cross-entropy loss across models and
      authors using only function words.} Follows the general format of Figure~\crossentropy~in the main text,
      but uses models trained on only function words. All content words are masked out using \texttt{<CONTENT>}.
      \textbf{A.} Average cross-entropy loss on
    \textit{Train}ing data and held-out test data from each author,
    plotted as a function of the number of training epochs. Each color
    denotes a model trained on a single author's work.  Error ribbons
    denote bootstrap-estimated 95\% confidence intervals over 10
    random seeds. \textbf{B.} Cross-entropy loss assigned to held-out
    test data by each author's model ($x$-axis). Held-out test data is
    either from the \textit{same} author (black) or from
    \textit{other} authors (gray). Each dot denotes the average loss
    (across all 1024-token chunks) for a single random seed.}
\label{fig:all-losses-function}
\end{figure*}

\begin{figure*}[h]
  \centering
  \includegraphics[width=\textwidth]{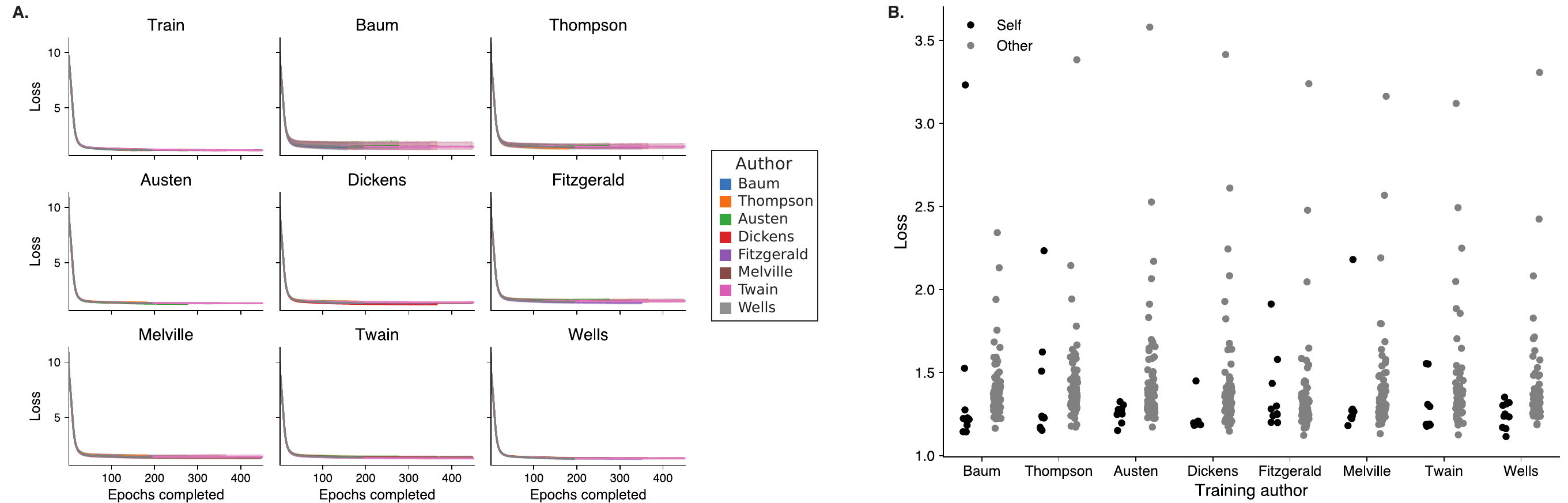}

  \caption{\textbf{Cross-entropy loss across models and
      authors using only parts of speech.} Follows the general format of Figure~\crossentropy~in the main text,
      but uses models trained on only parts of speech. All words are replaced with their corresponding part of speech tag.
      \textbf{A.} Average cross-entropy loss on
    \textit{Train}ing data and held-out test data from each author,
    plotted as a function of the number of training epochs. Each color
    denotes a model trained on a single author's work.  Error ribbons
    denote bootstrap-estimated 95\% confidence intervals over 10
    random seeds. \textbf{B.} Cross-entropy loss assigned to held-out
    test data by each author's model ($x$-axis). Held-out test data is
    either from the \textit{same} author (black) or from
    \textit{other} authors (gray). Each dot denotes the average loss
    (across all 1024-token chunks) for a single random seed.}
\label{fig:all-losses-pos}
\end{figure*}

\begin{figure*}[h]
  \centering
  \includegraphics[width=\textwidth]{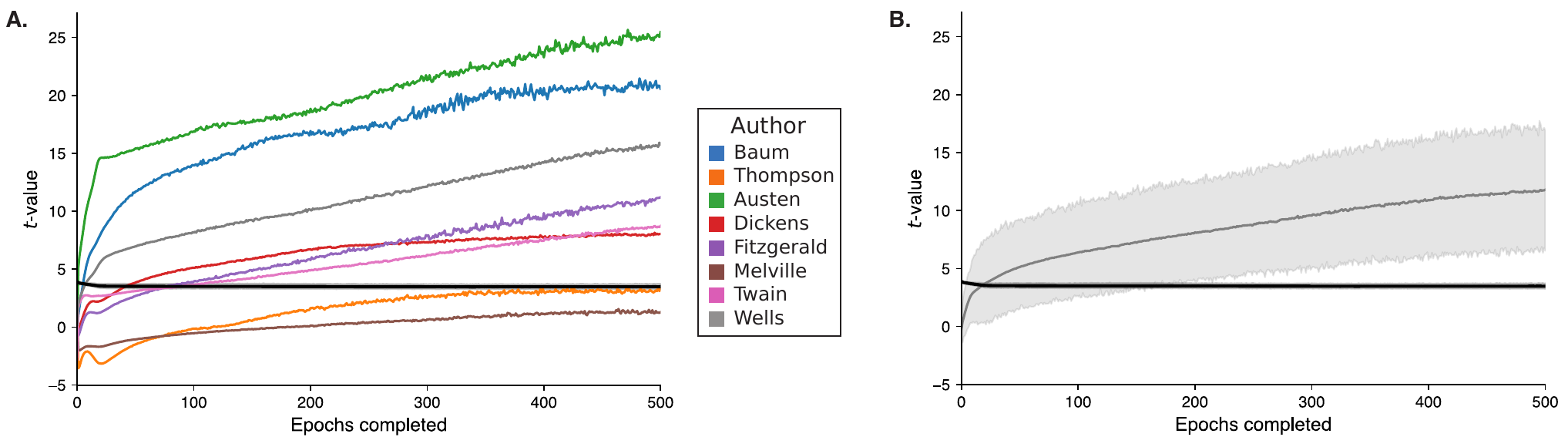}

\caption{\textbf{Same vs. other author comparisons, by model, using only
content words.} Follows the general format of Figure~\ttests~in the main text,
but uses models trained on only content words. All function words are masked
out using \texttt{<FUNC>}. \textbf{A.} Each curve denotes, as a function of the
number of training epochs, the the $t$-statistic from a $t$-test comparing the
distribution of losses (across random seeds) assigned to held-out texts from
the given author (color) versus held-out texts from all other authors.
\textbf{B.} The average $t$-statistic across all eight authors, as a function
of the number of training epochs. The black curves in both panels indicates the
average $t$-value corresponding to $p = 0.001$, for each epoch. Error ribbons
denote bootstrap-estimated 95\% confidence intervals across authors.}

\label{fig:t-stats-content}
\end{figure*}

\begin{figure*}[h]
  \centering
  \includegraphics[width=\textwidth]{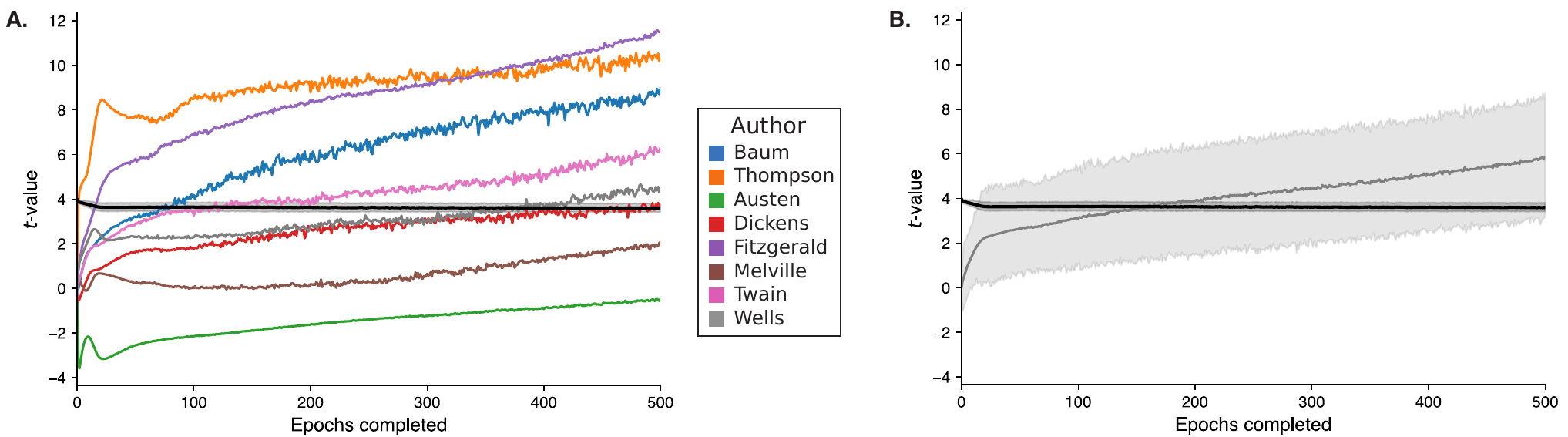}

\caption{\textbf{Same vs. other author comparisons, by model, using only
function words.} Follows the general format of Figure~\ttests~in the main text,
but uses models trained on only function words. All content words are masked
out using \texttt{<CONTENT>}. \textbf{A.} Each curve denotes, as a function of
the number of training epochs, the the $t$-statistic from a $t$-test comparing
the distribution of losses (across random seeds) assigned to held-out texts
from the given author (color) versus held-out texts from all other authors.
\textbf{B.} The average $t$-statistic across all eight authors, as a function
of the number of training epochs. The black curves in both panels indicates the
average $t$-value corresponding to $p = 0.001$, for each epoch. Error ribbons
denote bootstrap-estimated 95\% confidence intervals across authors.}

\label{fig:t-stats-function}
\end{figure*}

\begin{figure*}[h]
  \centering
  \includegraphics[width=\textwidth]{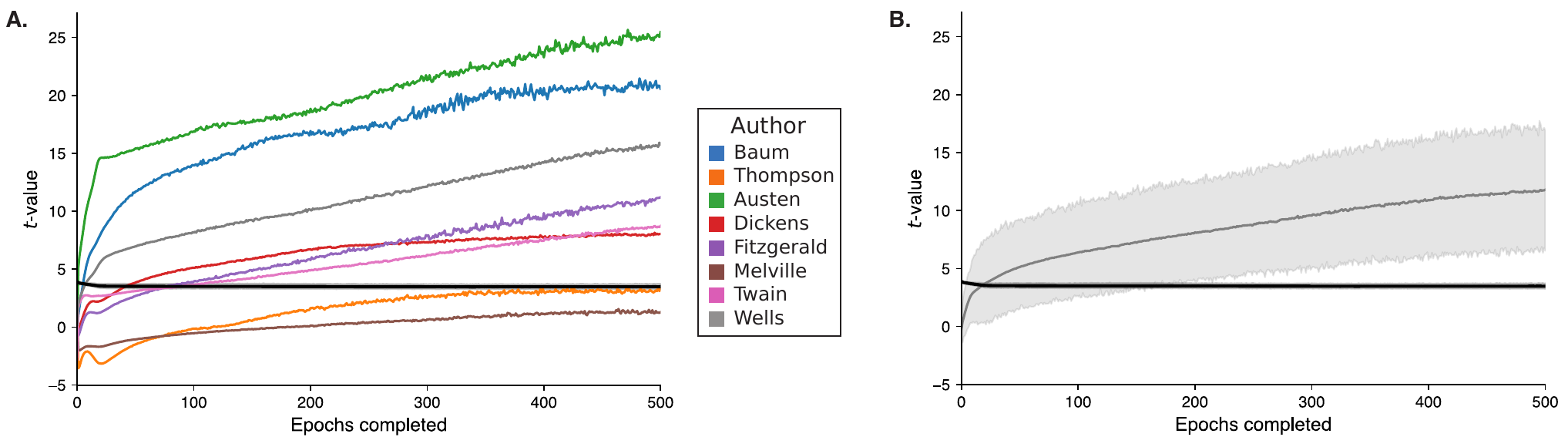}

\caption{\textbf{Same vs. other author comparisons, by model, using only parts
of speech.} Follows the general format of Figure~\ttests~in the main text, but
uses models trained on only parts of speech. All words are replaced with their
corresponding part of speech tag. \textbf{A.} Each curve denotes, as a function
of the number of training epochs, the the $t$-statistic from a $t$-test
comparing the distribution of losses (across random seeds) assigned to held-out
texts from the given author (color) versus held-out texts from all other
authors. \textbf{B.} The average $t$-statistic across all eight authors, as a
function of the number of training epochs. The black curves in both panels
indicates the average $t$-value corresponding to $p = 0.001$, for each epoch.
Error ribbons denote bootstrap-estimated 95\% confidence intervals across
authors.}

\label{fig:t-stats-pos}
\end{figure*}

\begin{figure*}[h]
  \centering
  \includegraphics[width=\textwidth]{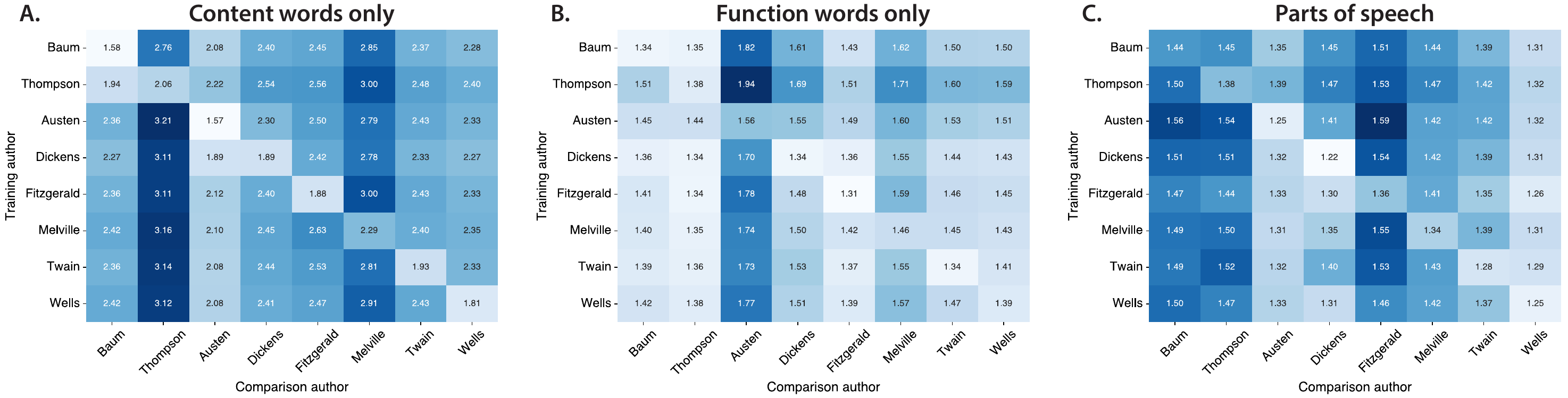}

\caption{\textbf{Confusion matrices.} Follows the general format of
Figure~\confusion~in the main text, but shows confusion matrices for models
trained on only content words (A), only function words (B), and only parts of
speech (C). Within each panel, the matrix displays the average cross-entropy loss assigned by
models trained on each author's writing (column) to held-out texts from each
author (row), after subtracting the native author's baseline loss.}
\label{fig:confusion-matrix-combined}

\end{figure*}

\begin{figure*}[t]
  \centering
  \includegraphics[width=\textwidth]{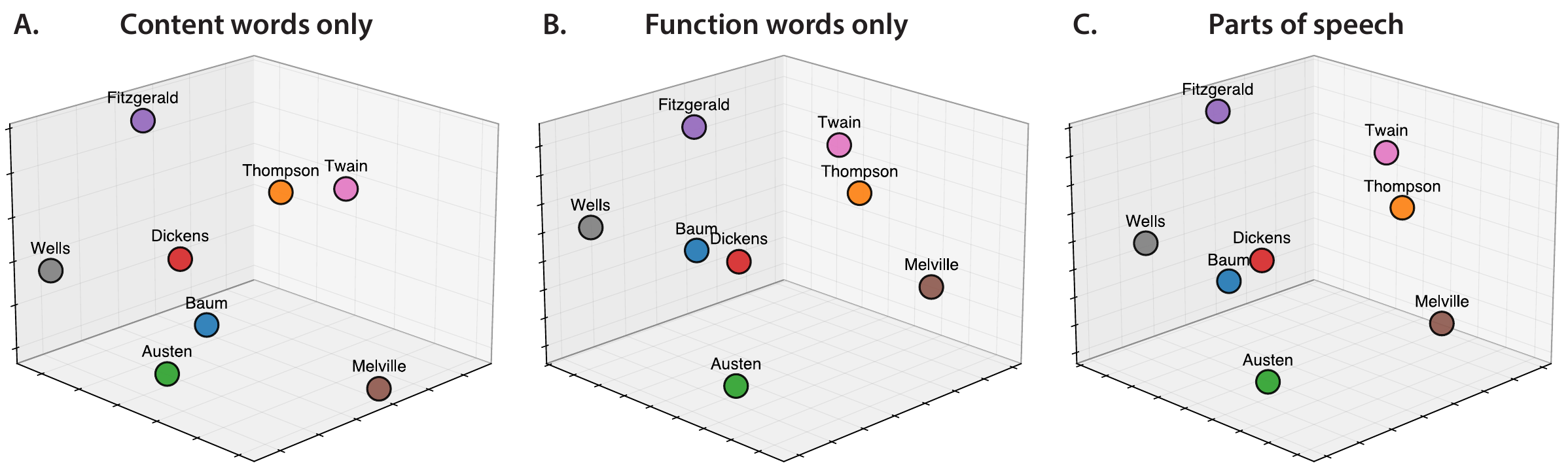}

\caption{\textbf{Multidimensional scaling plots.} Follows the general format of
Figure~\mds~in the main text, but shows MDS projections of the (symmetrized)
average cross entropy loss matrices shown in
Figure~\ref{fig:confusion-matrix-combined}, for models trained on only content
words (A), only function words (B), and only parts of speech (C).}
\label{fig:mds} \end{figure*}

\clearpage

\begin{table}[h]
\centering
\small
\begin{tabular}{lccc}
\hline
\textbf{Model} & \textbf{$t$-stat} & \textbf{df} & \textbf{$p$-value}\\
\hline
Baum         & 20.58 & 68.36 & $5.27 \times 10^{-31}$  \\
Thompson     & 3.29 & 11.33 & $6.97 \times 10^{-3}$  \\
Austen       & 25.46 & 70.33 & $3.20 \times 10^{-37}$  \\
Dickens      & 8.04 & 37.39 & $1.13 \times 10^{-9}$  \\
Fitzgerald   & 11.21 & 49.02 & $3.97 \times 10^{-15}$  \\
Melville     & 1.28 & 10.28 & $0.2274$  \\
Twain        & 8.73 & 22.50 & $1.12 \times 10^{-8}$  \\
Wells        & 15.79 & 71.87 & $4.53 \times 10^{-25}$  \\
\hline
\end{tabular}

\caption{\textbf{Loss differences between same-author and other-author texts
using only content words.} Follows the general format of Table~\authortable~in
the main text, but uses models trained on only content words. Each row displays
the results of a $t$-test comparing the average loss values assigned by each
author's model (after training is complete) to the author's held-out text and
to the other authors' randomly sampled texts.}

\label{tab:t-tests-content}
\end{table}

\begin{table}[h]
\centering
\small
\begin{tabular}{lccc}
\hline
\textbf{Model} & \textbf{$t$-stat} & \textbf{df} & \textbf{$p$-value}\\
\hline
Baum         & 8.97 & 20.85 & $1.34 \times 10^{-8}$  \\
Thompson     & 10.20 & 22.96 & $5.39 \times 10^{-10}$  \\
Austen       & -0.46 & 9.52 & $0.6581$  \\
Dickens      & 3.69 & 17.46 & $1.73 \times 10^{-3}$  \\
Fitzgerald   & 11.52 & 77.98 & $1.70 \times 10^{-18}$  \\
Melville     & 2.08 & 17.29 & $0.0529$  \\
Twain        & 6.31 & 34.66 & $3.14 \times 10^{-7}$  \\
Wells        & 4.49 & 15.94 & $3.76 \times 10^{-4}$  \\
\hline
\end{tabular}

\caption{\textbf{Loss differences between same-author and other-author texts
using only function words.} Follows the general format of Table~\authortable~in
the main text, but uses models trained on only function words. Each row displays
the results of a $t$-test comparing the average loss values assigned by each
author's model (after training is complete) to the author's held-out text and
to the other authors' randomly sampled texts.}

\label{tab:t-tests-function}
\end{table}

\begin{table}[h]
\centering
\small
\begin{tabular}{lccc}
\hline
\textbf{Model} & \textbf{$t$-stat} & \textbf{df} & \textbf{$p$-value}\\
\hline
Baum         & 0.04 & 9.22 & $0.9695$  \\
Thompson     & 1.05 & 10.91 & $0.3179$  \\
Austen       & 5.72 & 77.97 & $1.89 \times 10^{-7}$  \\
Dickens      & 4.41 & 62.85 & $4.14 \times 10^{-5}$  \\
Fitzgerald   & 0.43 & 14.25 & $0.6704$  \\
Melville     & 0.81 & 11.98 & $0.4337$  \\
Twain        & 2.43 & 20.88 & $0.0240$  \\
Wells        & 4.05 & 56.90 & $1.55 \times 10^{-4}$  \\
\hline
\end{tabular}

\caption{\textbf{Loss differences between same-author and other-author texts
using only parts of speech.} Follows the general format of Table~\authortable~in
the main text, but uses models trained on only parts of speech. Each row displays
the results of a $t$-test comparing the average loss values assigned by each
author's model (after training is complete) to the author's held-out text and
to the other authors' randomly sampled texts.}

\label{tab:t-tests-pos}
\end{table}